\documentclass{l4dc2025}

\usepackage{times}
\usepackage{amsmath}
\usepackage{amsfonts}
\usepackage{geometry}
\usepackage{enumitem}
\usepackage{hyperref}
\usepackage{xcolor}
\usepackage{booktabs}
\usepackage[utf8]{inputenc}
\usepackage[T1]{fontenc}
\usepackage{verbatim}
\usepackage{algorithm2e}
\usepackage{multirow}
\usepackage{multicol}
\usepackage{enumitem}
\usepackage{xcolor}
\usepackage{capt-of}
\usepackage{makecell}
\renewcommand{\cite}{\citep}
\usepackage[normalem]{ulem}
\graphicspath{{fig/}}
\title{Sparse-to-Field Reconstruction via\\ Stochastic Neural Dynamic Mode Decomposition}

\newif\ifshowresolved
\showresolvedtrue 
\ifshowresolved
    \newcommand{\resolved}[1]{\textcolor{gray}{[sd-resolved: #1]}}
    
    \newcommand{\yk}[1]{\textcolor{violet}{[yk: #1]}}
    \newcommand{\erase}[1]{\textcolor{black}{\sout{#1}}}
\else
    \newcommand{\resolved}[1]{}
    \newcommand{\yk}[1]{}
    \newcommand{\erase}[1]{}
    
\fi

\author{%
 \Name{Yujin Kim} \Email{yk826@cornell.edu}\\
 \Name{Sarah Dean} \Email{sdean@cornell.edu}\\
 \addr Department of Computer Science, Cornell University, Ithaca, NY, USA.
}

\begin{document}

\maketitle
\begin{abstract}
Many consequential real-world systems, like wind fields and ocean currents, are dynamic and hard to model. Learning their governing dynamics remains a central challenge in scientific machine learning. Dynamic Mode Decomposition (DMD) provides a simple, data-driven approximation, but practical use is limited by sparse/noisy observations from continuous fields, reliance on linear approximations, and the lack of principled uncertainty quantification. 
To address these issues, we introduce Stochastic NODE--DMD, a probabilistic extension of DMD that models continuous-time, nonlinear dynamics while remaining interpretable. 
Our approach enables continuous spatiotemporal reconstruction at arbitrary coordinates and quantifies predictive uncertainty. 
Across four benchmarks, a synthetic setting and three physics-based flows, it surpasses a baseline in reconstruction accuracy when trained from only 10\% observation density. 
It further recovers the dynamical structure by aligning learned modes and continuous-time eigenvalues with ground truth.
Finally, on datasets with multiple realizations, our method learns a calibrated distribution over latent dynamics that preserves ensemble variability rather than averaging across regimes. 
Our code is available \href{https://github.com/sdean-group/Stochastic-NODE-DMD}{here}.
\end{abstract}

\section{Introduction}
\label{sec:intro}

Data-driven identification of high-dimensional systems—inferring dynamics from sequential data—is a core challenge in scientific machine learning~\cite{brunton2016discovering,raissi2019physics,lusch2018deep}. Applications span large-scale phenomena such as weather and climate~\cite{shi2015convolutional,rasp2020weatherbench,pathak2023climate} and finer-scale neuroscience problems like brain imaging reconstruction~\cite{lee2021multitask,mechelli2005voxel,misra2025brain}. Beyond scientific value, system identification underpins planning and control in open environments: wind-field models aid balloon navigation and UAV control~\cite{bellemare2020autonomous,sydney2013dynamic}, ocean-current models guide marine robots~\cite{vasilijevic2017coordinated,wiggert2022navigating}, and traffic-flow models inform autonomous driving~\cite{wei2014autonomous}. In these applications, probabilistic modeling improves robustness and long-horizon stability.

Existing approaches fall into three groups: physics-based, operator-theoretic (Koopman-based), and black-box. Physics-based models~\cite{brunton2016discovering,raissi2019physics} leverage governing equations and are interpretable, but can be noise-sensitive, expensive at scale, and rarely probabilistic. Operator methods grounded in spectral/Koopman theory~\cite{Mezic2005}—including DMD~\cite{Rowley2009,Schmid2010,tu2014exact,lusch2018deep} and neural operators~\cite{lu2019deeponet,li2020fourier,kovachki2023neural}—yield interpretable linear surrogates~\cite{Kutz2016} but are constrained by linear evolution and grid-based operations, limiting performance with sparse observations and strong nonlinearities. Hybrid efforts that pair neural implicit representations~\cite{mildenhall2021nerf,niemeyer2022regnerf,muller2022instant} with optimized DMD~\cite{askham2018variable,sashidhar2022bagging} (e.g., \cite{saraertoosi2025neural}) expand expressivity yet remain fundamentally linear and deterministic. Black-box sequential models \cite{chen2018,pathak2018model,kidger2020neural} scale well and are resolution-agnostic~\cite{brunton2020machine}, but often lack uncertainty quantification, struggle to generalize, destabilize over long horizons, and may violate physics~\cite{chen2018,kidger2020neural,brunton2020machine}. Probabilistic variants~\cite{krishnan2015deep,doerr2018probabilistic,alvarez2013linear} add uncertainty quantification but demand more data and computation and reduce interpretability.

In many dynamical systems, stochasticity reflects both imperfect observation and unmodeled dynamics (e.g., measurement noise, environmental perturbations). Moreover, practical constraints—high experimental costs, physical inaccessibility, and sensor limitations—lead to sparse observations. 
Consequently, a method that is \uline{(i) physically interpretable}, provides \uline{(ii) uncertainty quantification}, and learns effectively from \uline{(iii) sparse, noisy datasets} is needed.
This paper introduces Stochastic Neural Ordinary Differential Equation Dynamic Mode Decomposition (Stochastic NODE-DMD), a deep probabilistic model for system identification using DMD. 
We focus on stochastic system identification from sparse and nonlinear datasets.

Prior work has developed probabilistic frameworks for DMD~\cite{Takeishi2017} and neural implicit representations for spatial bases~\cite{saraertoosi2025neural}.
However, neither addresses the joint challenges of continuous spatiotemporal reconstruction with temporally coherent uncertainty quantification from sparse observations.
Our method advances beyond these approaches by (1) replacing fixed spatial bases with an implicit neural encoder that enables continuous reconstruction from arbitrary sparse measurements, and (2) introducing a stochastic Neural ODE~\cite{chen2018,tzen2019,li2020scalable} for latent dynamics that maintains long-term temporal coherence.
This unified framework provides uncertainty quantification 
for both measurement noise and dynamics variance across multiple realizations.

We evaluate Stochastic NODE-DMD on four datasets: a synthetic dataset with linear mode evolution and added noise, and three physics-based benchmarks including the Gray-Scott reaction-diffusion model for chemical dynamics~\cite{pearson1993complex}, Navier-Stokes equations for modeling vorticity flow, and flow past a cylinder~\cite{brunton2020machine}. Using only 10\% of fixed datapoints across sequences to simulate sparsity, we assess reconstruction quality and dynamics retrieval. Comparisons of learned modes and coefficients against ground-truth parameters demonstrate the model's reliability, while tests on datasets with multiple dynamics highlight effective uncertainty quantification.

\section{Background and Problem Formulation}
\subsection{Problem setting}
Consider a continuous-time nonlinear dynamical system with state $\mathbf{x}(t)$ governed by dynamics $f$: 
\begin{equation}
\label{eq:dynamics}
\dot{\mathbf{x}}(t) = f(\mathbf{x}(t)), \quad \mathbf{x}(t) \in \mathcal X,    
\end{equation}
In many applications, such as fluid dynamics or wave propagation, the overall state $\mathbf{x}(t)$ is as a continuous function over a spatial domain.
With spatial coordinate \(\mathbf{s} \in \Omega \subset \mathbb{R}^d\), \(\mathbf{x}(\mathbf{s}, t) \in \mathbb{C}\) 
is the value of the state at coordinate \(\mathbf{s}\) and time \(t\), and \(\mathcal{X} = L^2(\Omega)\) is the function space of square-integrable fields over the spatial domain \(\Omega\). As a result, Eq.~\eqref{eq:dynamics} typically manifests as a partial differential equation (PDE), though it is commonly approximated by discretization for numerical simulation.
In particular, a grid of $\Omega$ is defined, denoted $S_\Omega\subset \Omega$ with $|S_\Omega|=n$.
Then the system is approximated with a finite dimensional state in $\mathbb C^n$ and corresponding approximate dynamics. For an $\epsilon$ grid, we typically require $n$ to scale like $\mathrm{vol}(\Omega)/\epsilon^d$, so even in this finite dimensional approximation, the state is high dimensional.

We instead consider observations that are spatially sparse, such as measurements from a few fixed meteorological or oceanographic stations.
Define a finite set of spatial coordinates \(S = \{\mathbf{s}_1, \dots, \mathbf{s}_m\} \subset \Omega\)
and let $\mathbf x(S,t)\in\mathbb C^m$ denote the values of the state at those coordinates at time $t$.
We observe noisy measurements at discrete times \(t_k\) (\(k = 0, \dots, p-1\)) as $\mathbf{y}_k = \mathbf x(S, t_k) + {\eta}_k$, where \({\eta}_k \in \mathbb{C}^m\) represents additive measurement noise.
The resulting dataset consists of the observation sequence $\{\mathbf{y}_k\}_{k=0}^{p-1}$, such as time-indexed sensor readings.
The goal is to learn a low-dimensional and interpretable model that captures the underlying spatiotemporal dynamics, enables prediction of fields $\mathbf x(\mathbf{s}, t)$ at arbitrary coordinates $\mathbf{s}$ (including unseen full-resolution grids), and quantifies uncertainty due to factors like noise, subsampling, and model mismatch. 
To this end, we formulate the problem as inferring a stochastic dynamic mode decomposition over continuous space by modeling the mode coefficient as a latent state.



\subsection{Linear Evolution and Dynamic Mode Decomposition}
\label{sec:linear_evolution}
While the governing dynamics in Eq.~\eqref{eq:dynamics} are nonlinear and infinite-dimensional, 
many phenomena of interest can be well represented in a low dimensional manner.
This is achieved by projecting the spatiotemporal data onto a set of spatial modes \(W_i\), yielding time-dependent mode coefficients \(\phi_{i}\):
\begin{equation}
\label{eq:modal_decomp}
\mathbf{x}(\mathbf s, t) \approx \sum_{i=1}^r \phi_{i}(t) W_i(\mathbf s),
\quad \mathbf{y}_k \approx \sum_{i=1}^r \phi_{i,k} \mathbf{w}_i + {\eta}_k,
\end{equation}
where modes \(\mathbf{w}_i = W_i(S) \in \mathbb{C}^m\) have  \(\phi_{i,k}\in \mathbb{C}\) representing their temporal amplitude at time \(t_k\)
and $r$ is the number\footnote{In standard DMD, the maximum number of computable modes is limited by the rank of the data matrix. However, it is common to truncate to a smaller $r$ by retaining only the dominant singular values from the SVD step, as this reduced-rank approximation is sufficient to capture the essential dynamics in many systems of interest \cite{tu2014exact,schmid2010dmd}.}
 of modes. 
Under an additional assumption of linear mode evolution, the coefficients evolve as:
\begin{equation}
\label{eq:linear_evolution}
\dot \phi_i(t) = \lambda_i \phi_i(t),\quad 
\phi_{i,k+1} = e^{\lambda_i \Delta t} \phi_{i,k},
\end{equation}
with complex eigenvalue \(\lambda_i = \alpha_i + j \beta_i\) encoding growth rate \(\alpha_i\) and frequency \(\beta_i\).  
This linear structure, while an approximation, enables predictive modeling and spectral analysis of complex systems~\cite{Rowley2009,Schmid2010}.
This approximation is theoretically grounded in Koopman operator theory~\cite{Mezic2005}, which lifts nonlinear dynamics into a linear (infinite-dimensional) space of observables. 

Dynamic Mode Decomposition (DMD) is a tractable approach to linear dynamics on high or infinite dimensional states.
Given an observation $\mathbf{y}_k \in \mathbb{C}^m$ at times $t_k$, collection of sequential observation data pair $Y$ and $Y'$ are defined as $Y = [\mathbf{y}_0, \mathbf{y}_1, \dots, \mathbf{y}_{p-1}] \in \mathbb{C}^{m \times p}$ and $Y' = [\mathbf{y}_1, \mathbf{y}_2, \dots, \mathbf{y}_p] \in \mathbb{C}^{m \times p}$.
DMD computes a matrix $ A $ such that $ Y' \approx A Y $, from which the eigenvalues $ \mu_i $ and mode matrix $ W \in \mathbb{C}^{m \times r} $ (with $ r $ effective modes) are obtained. 
The discrete eigenvalues satisfy $ \mu_i \approx e^{\lambda_i \Delta t} $, linking DMD to the continuous linear evolution in Eq.~\eqref{eq:linear_evolution}.
In practice, operating on vectorized, fixed-grid snapshots provides computational simplicity but reduces flexibility on irregular or continuous spatial domains; the linear-evolution assumption yields clear mode structure yet can leave nonlinear residuals unmodeled in complex systems; and the standard deterministic formulation emphasizes point estimates over calibrated uncertainty, which can be challenging under sparse sensors, noise, or missing data.

\subsection{Probabilistic DMD as a Generative Model}
\label{sec:prob_dmd}
To incorporate uncertainty and enable principled probabilistic inference, DMD can be reformulated as a latent-variable generative model~\cite{Takeishi2017}.  
In this framework, the mode coefficients from the linear evolution (Section~\ref{sec:linear_evolution}) are treated as latent variables \(\phi_{i,k} \in \mathbb{C}\).
For each \emph{pair} of consecutive snapshots \((\mathbf{y}_k, \mathbf{y}_{k+1})\), the conditional likelihoods are defined as:
\begin{align}
\label{eq:gen_model}
\mathbf{y}_k \mid \{\phi_{i,k}\}_{i=1}^r
  \sim \mathcal{CN}\left( \sum_{i=1}^r \phi_{i,k} \mathbf{w}_i, \sigma^2 I_m \right), \quad
\mathbf{y}_{k+1} \mid \{\phi_{i,k}\}_{i=1}^r
  \sim \mathcal{CN}\left( \sum_{i=1}^r e^{\lambda_i \Delta t} \phi_{i,k} \mathbf{w}_i, \sigma^2 I_m \right),
\end{align}
with standard complex Gaussian priors \(\phi_{i,k} \sim \mathcal{CN}(0,1)\). Maximizing the marginal likelihood recovers classical DMD in the noise-free limit.  
However, because the generative model is defined \emph{pairwise}—with independent latent states \(\boldsymbol{\phi}_k\) for each transition—the global linear evolution in Eq.~\eqref{eq:linear_evolution} is not enforced across time steps.  
This breaks long-term dynamical consistency, potentially leading to accumulating errors in multi-step forecasting and unstable mode amplitude trajectories under noise or missing data.

\subsection{Nonlinear State-Space Formulation for Stochastic NODE--DMD} 
We extend the probabilistic generative model of Bayesian DMD~\cite{Takeishi2017} into a nonlinear state-space framework supporting continuous-time evolution and uncertainty quantification. Let $\boldsymbol{\phi}_k \in \mathbb{C}^r$ be the low-dimensional latent state vector encoding $r$ modes (with $r \ll m$), $\Lambda = \text{diag}(\lambda_1, \dots, \lambda_r) \in \mathbb{C}^{r \times r}$ the diagonal matrix of eigenvalues, and $W \in \mathbb{C}^{m \times r}$ the mode matrix. Then:
\begin{align}
\boldsymbol{\phi}_{k+1} &= e^{\Lambda \Delta t} \boldsymbol{\phi}_k + f_\theta(\boldsymbol{\phi}_k, t_k) \Delta t + \zeta_k, \quad
\zeta_k \sim \mathcal{CN}(\mathbf{0}, \tau^2 \Delta t \cdot I_r), \\
\mathbf{y}_k &= W \boldsymbol{\phi}_k + \mathbf{\eta}_k, \quad
\mathbf{\eta}_k \sim \mathcal{CN}(\mathbf{0}, \sigma^2 I_m),
\end{align} 
where $f_\theta$ is a neural network capturing nonlinear residual dynamics (parameterized by $\theta$), $\zeta_k$ represents process noise, and $\mathbf{\eta}_k$ denotes observation noise.

To derive a differentiable continuous-time model, set \(t_k = k \Delta t\) and take the limit \(\Delta t \to 0\). The discrete increment \({\zeta}_k = \tau \cdot \Delta B_k\) with \(\Delta B_k \sim \mathcal{CN}(\mathbf{0}, \Delta t \cdot I_r)\) converges in distribution to a Brownian increment, yielding the stochastic differential equation:
\begin{equation}
    d\boldsymbol{\phi}_t = \bigl( \Lambda \boldsymbol{\phi}_t + f_\theta(\boldsymbol{\phi}_t, t) \bigr) dt + \tau \, dB_t.
    \label{eq:sde}
\end{equation}
Equation~\eqref{eq:sde} thus defines a neural stochastic differential equation combining linear evolution drift, nonlinear correction, and process noise.  $\Lambda \boldsymbol{\phi}_t$ corresponds to the classical linear evolution, while the neural residual $ f_\theta $ provides adaptive corrections for nonlinearity and model mismatch.  
In the limit $ f_\theta \to 0 $ and $ \tau^2 \to 0 $, the formulation reduces to deterministic DMD.

\section{Stochastic Neural Ordinary Differential Equation DMD (NODE--DMD)}
\label{sec:method}
\begin{figure}
    \centering
    \includegraphics[width=\linewidth]{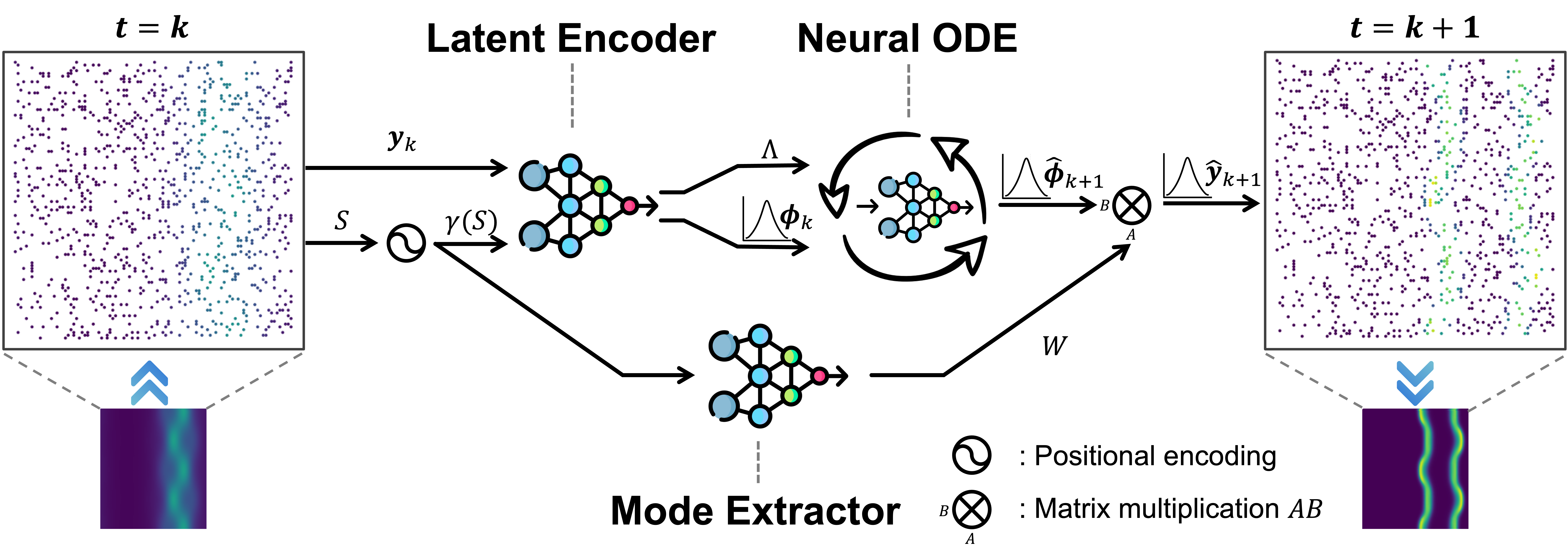}
    \caption{\textbf{Overview of the Stochastic NODE--DMD architecture.} At time step $k$, subsampled measurements $\mathbf{y}_k$ and set of spatial coordinates $S$ are fed into the \texttt{Latent Encoder}, which outputs eigenvalues $\Lambda$ and the latent distribution $p(\boldsymbol{\phi}_k)$ . The \texttt{Neural ODE} evolves this distribution forward to time $k+1$. The predicted latent state distribution $p(\boldsymbol{\hat{\phi}}_{k+1})$ is then combined with spatial modes from the \texttt{Mode Extractor} $W$ to reconstruct the distribution of subsampled field at time $k+1$, enabling uncertainty-aware, grid-free forecasting.}
    \label{fig:architecture}
\end{figure}

We extend the probabilistic DMD framework proposed by~\citet{Takeishi2017}, by incorporating continuous-time nonlinear latent evolution via Neural ODEs with stochastic diffusion~\cite{chen2018,tzen2019}.
We also build upon NeuralDMD (NDMD)~\cite{saraertoosi2025neural} which integrates neural implicit representation for sparse reconstruction.
Our proposed method, termed Stochastic Neural ODE DMD (NODE-DMD), learns uncertainty-aware dynamics within a generative modeling framework. It combines the linear spectral structure of DMD with data-driven nonlinear corrections, and supports sparse, noisy observations.
Figure~\ref{fig:architecture} illustrates the overall architecture of Stochastic NODE--DMD.  
Given subsampled measurements \(\mathbf{y}_k\) at fixed spatial coordinates \(S\), the model predicts the state distribution at time \(k+1\).  
Inputs may originate from the dataset (``teacher forcing'' with \(\mathbf{y}_k\)) or prior model predictions (autoregressive forecasting from \(\mathbf{\hat{y}}_k\)).


\label{sec:overall_structure}

There are three main components: the \texttt{Mode Extractor} outputs spatial mode functions,
and the \texttt{Latent Encoder} 
outputs eigenvalues \(\Lambda\) and the initial latent distribution \(\mathcal{CN}(\boldsymbol{\mu}_{\phi_k}, \Sigma_{\phi_k})\).  
Both depend on a positional encoding $\gamma(\mathbf s)$.
The stochastic \texttt{Neural ODE} block evolves this distribution forward in continuous time.  
The predicted latent state distribution \(p(\boldsymbol{\hat{\phi}}_{k+1}) = \mathcal{CN}(\boldsymbol{\mu}_{\phi_{k+1}}, \Sigma_{\phi_{k+1}})\) is combined with 
the mode function \(W_\psi(\gamma(\mathbf s))\) to generate field prediction at arbitrary coordinates
\(
\hat{\mathbf y}(\mathbf{s}, t_{k+1}) = \sum_{i=1}^r \hat{\boldsymbol{\phi}}_{k+1, i} \, [W_\psi(\gamma(\mathbf{s}))]_i
\).
By using 
\(\mathbf{\hat{y}}_{k+1}=\hat{\mathbf y}(S, t_{k+1})\), this architecture enables uncertainty-aware, multi-step forecasting from sparse and noisy observations. 

\subsection{Mode Extractor and Latent Encoder with Positional Encoding}
\label{sec:mode_extractor_latent_encoder}

To enable reconstruction at arbitrary coordinates $\mathbf{s} \in \Omega$, we parameterize spatial modes using a neural implicit representation $W_\psi$~\cite{mildenhall2021nerf,muller2022instant}, parameterized by $\psi$.
To effectively encode raw spatial coordinates and enable the network to capture high-frequency details in continuous domains, we apply a positional encoding $\gamma: \mathbb{R}^d \to \mathbb{R}^{2L d}$ (where $d$ is the coordinate dimensionality and $L$ is the number of frequency bands) to individual coordinates before feeding them into the relevant modules~\cite{tancik2020fourier,mildenhall2021nerf}. This encoding transforms each coordinate dimension $s_j$ as
$\gamma(s_j) = \left[ s_j, \sin(2^0 \pi s_j), \cos(2^0 \pi s_j), \dots, \sin(2^{L-1} \pi s_j), \cos(2^{L-1} \pi s_j) \right]$.
We compute the encoded set $\gamma(S) = \{\gamma(\mathbf{s}_i)\}_{i=1}^m$, which is used as input to the \texttt{Mode Extractor}.
The pair $(\mathbf{y}_k, \gamma(S))$ is used as input to the
\texttt{Latent Encoder}
which determines the mode coefficient $\boldsymbol{\phi}$ as probabilistic latent state using variational inference encoding~\cite{kingma2013auto}.


\subsection{Stochastic Neural ODE for Latent Evolution}
\label{sec:stochastic_node}

The stochastic \texttt{Neural ODE} enforces linear evolution across time steps while allowing nonlinear corrections and intrinsic uncertainty propagation.
It evolves the latent mode state $\boldsymbol{\phi}(t)\in\mathbb{C}^r$ continuously via the SDE in Eq.~\eqref{eq:sde}, where the linear drift is encoded by $\Lambda$, the nonlinear residual is modeled by $f_\theta$, and the diffusion term $\tau\,d\mathbf{B}_t$ captures intrinsic uncertainty. 
To overcome the independent pairwise-transition limitation of the generative model in Bayesian DMD~\cite{Takeishi2017}, we propagate the {distribution} of $\boldsymbol{\phi}$ with a stochastic \texttt{Neural ODE}, allowing data-driven corrections and uncertainty transport. Furthermore, we reinforce temporal continuity using the consistency loss in Sec.~\ref{sec:training_objective}. 
We discretize the SDE over substeps of $\delta t$ with an uncertainty-aware Euler--Maruyama scheme~\cite{kloeden1977numerical}.
Implementation details, including multi-substep updates and covariance propagation, are given in Appendix~\ref{app:snode_details}.

\subsection{Training Objective and Rollout Strategy}
\label{sec:training_objective}
The model is trained end-to-end using a composite loss function that balances reconstruction accuracy, latent regularization, and temporal consistency. (Full mathematical details are provided in Appendix~\ref{app:loss}). 
For a transition from time \(t_k\) to \(t_{k+1}\), let \(\mathbf{y}_{k+1} \in \mathbb{C}^m\) be the target observation, \((\boldsymbol{\hat\mu}_y, \boldsymbol{\hat\sigma}_y^2)\) the predicted mean and variance of the reconstructed field, and \((\boldsymbol{\mu}_{\phi}, \boldsymbol{\sigma}_{\phi}^2)\), \((\boldsymbol{\hat\mu}_{{\phi}}, \boldsymbol{\hat\sigma}_{{\phi}}^2)\) the latent distributions from the \texttt{Latent Encoder} and Stochastic \texttt{Neural ODE}, respectively.
\begin{itemize}[noitemsep,topsep=0pt,parsep=0pt,partopsep=0pt,leftmargin=* ]
    \item \textbf{Reconstruction Loss}: A Gaussian negative log-likelihood from $p({\hat {\mathbf y}_{k+1}})=(\boldsymbol{\hat\mu}_y, \boldsymbol{\hat\sigma}_y^2)$ to $\mathbf y_{k+1}$ encourages accurate field prediction while adapting to data uncertainty.
    \item \textbf{Latent KL Regularization}: A KL divergence term regularizes the latent mode coefficients \(\boldsymbol{\hat\phi}_i\) toward a standard complex Gaussian prior.
    \item \textbf{Consistency Loss}: The latent distribution matching loss aligns the encoder-estimated distribution \(p(\boldsymbol{\phi}_{k+1} \mid \mathbf{y}_{k+1}, \gamma(S))\) with the SDE-propagated distribution \(p(\boldsymbol{\hat{\phi}}_{k+1} \mid \boldsymbol{\phi}_{k})\) using MSE and KL-divergence. 
\end{itemize}
The total loss is a weighted sum of these three loss terms. 
We train the model end-to-end with a curriculum that gradually transitions inputs from ground truth ($\mathbf{y}_k$) to model predictions ($\hat{\mathbf{y}}_k$), stabilizing training and enabling robust long-horizon autoregressive forecasts; 
details are provided in Appendix~\ref{app:train_detail}.

\section{Experiments}
\paragraph{Models.} We evaluated Stochastic NODE--DMD on four datasets and benchmarked it against Neural DMD (NDMD) \citep{saraertoosi2025neural}, which learns DMD components via a neural implicit representation. For a fair comparison, we set the mode rank to \(r=4\) for Synthetic Sequence task and \(r=8\) for all other tasks, and aligned key training hyperparameters such as the number of epochs and batch size across methods.

\begin{figure}[t!]
    \centering
    \includegraphics[width=0.9\linewidth]{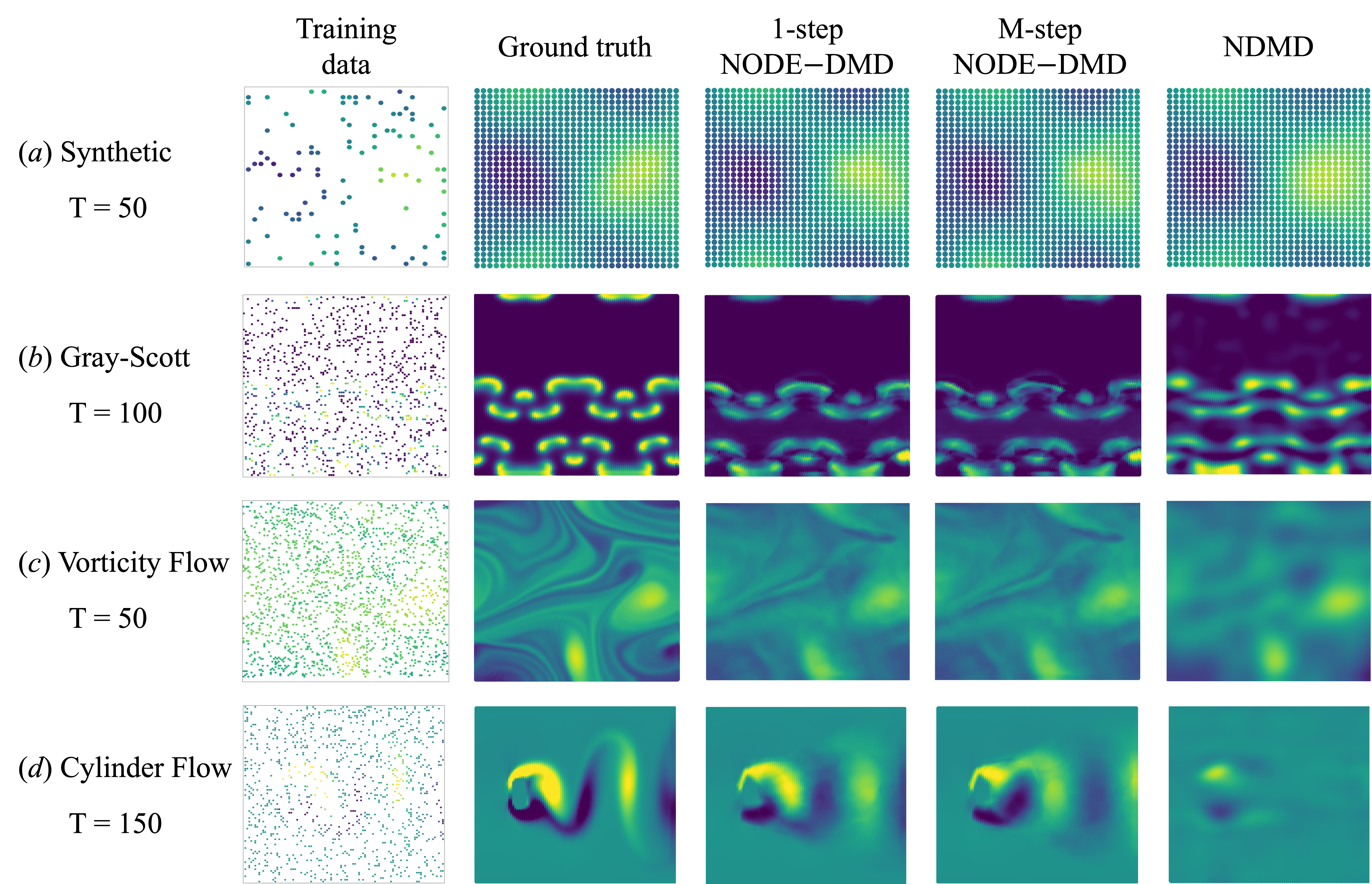}
    \caption{\textbf{Qualitative reconstruction results from sparse observations.}
    Row labels list the task and rollout length $T$ (final time step).  Columns: (1) training data—10\% of fixed spatial points sampled from the ground truth, (2) ground-truth full field, (3) NODE--DMD 1-step (teacher-forced), (4) NODE--DMD $m$-step (autoregressive), and (5) NDMD $m$-step.
    }
    \label{fig:reconstructions}
\end{figure}

\paragraph{Simulated dynamical systems.}
We evaluate Stochastic NODE--DMD on four simulated systems, ranging from a linear synthetic sequence with known ground truth to nonlinear PDE-governed fluid flows. In Fig.~\ref{fig:reconstructions}, the second column shows the ground-truth field for each task. All systems are discretized on structured grids and evolved for $T=50$--$150$ time steps to capture salient dynamics and to enable both short- and long-horizon evaluation. In the sparse-observation experiments (Sec.~\ref{sec:reconstructions} and~\ref{sec:gt-dynamics}), we train with 10\% fixed spatial sensors to emulate sparse measurements and evaluate reconstructions on the full grid.
\begin{itemize}[noitemsep,topsep=0pt,parsep=0pt,partopsep=0pt,leftmargin=* ]
    \item \textbf{Synthetic Sequence}: Four predefined spatial modes with exponential temporal evolution (Eq.~\eqref{eq:modal_decomp}) and additive observation noise. This mirrors common synthetic setups used to validate DMD/Koopman methods~\cite{dawson2016characterizing,askham2018variable,zhang2017evaluating} preserving the ground-truth modes and eigenvalues. The dataset uses a $32\times32$ grid and sequence length $T=50$ (Fig.~\ref{fig:reconstructions}(a)).
    \item \textbf{Gray-Scott Reaction-Diffusion}: A coupled PDE system modeling two reacting and diffusing chemical species~\cite{pearson1993complex}. The dataset uses a $100\times 100$ grid and sequence length $T=100$ (Fig.~\ref{fig:reconstructions}(b)). 
    \item \textbf{2D Navier-Stokes Spectral Vorticity}: Incompressible 2D flow in vorticity using \texttt{torch-cfd}~\cite{cao2024spectral}. The dataset uses a $100\times 100$ grid and sequence length $T=50$  (Fig.~\ref{fig:reconstructions}(c)).
    \item \textbf{2D Flow Past a Cylinder}: Viscous incompressible flow around a circular cylinder, governed by the Navier-Stokes equations simulated using \texttt{torch-cfd}~\cite{cao2024spectral} with finite volume method and pressure projection. The dataset uses a $128\times 128$ grid and sequence length $T=150$. (Fig.~\ref{fig:reconstructions}(d)).
\end{itemize}
Details of parameters, initial conditions, numerical schemes, and implementation notes are provided in Appendix~\ref{app:simulations}.
\subsection{Reconstructing continuous spatiotemporal dynamics from sparse observations.}\label{sec:reconstructions}
We evaluate whether NODE--DMD can infer missing spatiotemporal information from sparse observations while remaining reliable to the underlying dynamics. To this end, only 10\% of spatial locations are used as fixed sensors and observed over the entire sequence.

\paragraph{Qualitative Results.} 
Figure~\ref{fig:reconstructions} visualizes the final rollout segment for each task. For NODE--DMD we report both 1-step (teacher-forced) and $m$-step (autoregressive) settings, whereas NDMD is evaluated in $m$-step.  With only sparse measurements, NODE--DMD preserves coherent pattern and sharp phase alignment, whereas NDMD produces blurred reconstructions that lose fine-scale structures.
The Cylinder Flow task highlights characteristic behaviors: 1-step NODE--DMD captures sharp geometric detail and phase alignment, whereas in \(m\)-step prediction, it maintains plausible geometry but exhibits phase lag due to accumulated frequency misalignment in long-horizon prediction. 
By contrast, Neural DMD tends to average out the vortices shedding. 
Learning spatial modes as continuous functions of coordinates enables interpolation at unobserved locations and effectively super-resolves missing regions, mitigating the ambiguity induced by fixed sparse sensors.

\paragraph{Quantitative results.} Table~\ref{tab:reconstructions} reports the time-averaged per-pixel $L_1$ error. On the near-linear, mode-separable Synthetic DMD Sequence task, NODE--DMD is comparable to NDMD. On the more nonlinear systems—Gray-Scott, Spectral Vorticity, and Cylinder Flow—NODE–DMD achieves consistently lower 1-step error than NDMD, and shows comparable error in longer m-step forecasts.

As expected, $m$-step rollouts generally underperform 1-step because autoregressive recovery proceeds open-loop from the first frame, so small modeling errors accumulate over time. On Cylinder, periodicity makes pointwise $L_1$ highly phase-sensitive: small frequency bias causes phase drift and is penalized without time-shift alignment, increasing the $m$-step error. Note that the smaller error achieved by NDMD comes at the cost of ``averaging out'' uncertainty in the phase as shown in Fig.~\ref{fig:reconstructions}. We further analyze the role of stochastic variance in Section~\ref{sec:distributions}.
\begin{table}
\small
\centering
\begin{tabular}{c|cc|cc|cc|cc}
\hline
\multirow{1}{*}{System} &
\multicolumn{2}{c|}{\textsc{DMD sequence}} &
\multicolumn{2}{c|}{\textsc{Gray--Scott}} &
\multicolumn{2}{c|}{\textsc{Vorticity Flow}} &
\multicolumn{2}{c}{\textsc{Cylinder Flow}} \\

\cline{1-9}
{horizon} & 1-step & m-step & 1-step & m-step & 1-step & m-step & 1-step & m-step \\
\hline
\hline
NDMD & n/a &0.0442 & n/a & 0.0049 & n/a & 0.0047 & n/a & 0.0467\\
NODE--DMD & 0.0466 &0.0455 & 0.0046 & 0.0076 & 0.0024 & 0.0024 & 0.0067 & 0.0473\\
\hline
\end{tabular}
\caption{$L_1$ error of full snapshot reconstruction from sparse (10\%) observations}
\label{tab:reconstructions}
\end{table}

\subsection{Recovering Ground-Truth Dynamics from Noisy, Sparse Observations}
\label{sec:gt-dynamics}

We test whether NODE--DMD recovers continuous-time dynamics from noisy sequences observed at fixed $10\%$ sensors. 
Additional details are presented in Appendix~\ref{app:dyanmics_retriever}; here we report qualitative results.

Fig.~\ref{fig:mode_contour} visualizes mode portraits at a random time step ($t{=}14$).
For each mode $k$, we draw isocontours of the instantaneous spatial contribution
$\lvert W_k(x)\,\phi_k(t)\rvert$.
To ensure comparability across time and between GT/predictions, we fix absolute contour levels per mode using the GT sequence: pool $\lvert W_k(x)\phi_k(t)\rvert$ over all $(x,t)$ and set levels to the $(30,60,90)$-th percentiles of that pooled distribution.\footnote{A spatially constant GT mode (e.g., mode 1) yields nearly flat $\lvert W_k\rvert$ and degenerate isocontours; such cases may be omitted as in Fig.~\ref{fig:mode_contour}(a).}
Portraits in GT mode (left side in Fig.~\ref{fig:mode_contour}) and portraits predicted by the model (right side in Fig.~\ref{fig:mode_contour})
exhibit matched high-contribution regions, indicating that the model organizes spatial energy consistently with the ground-truth dynamics.
\subsection{Learning a distribution over dynamics across realizations}\label{sec:distributions}
We test whether NODE--DMD learns a posterior over spatiotemporal fields that captures variability across realizations of the same system (identical initial condition, different dynamics parameters). We train two models on the Vorticity Flow benchmark at a lower resolution than in Section~\ref{sec:reconstructions}. To isolate dynamics uncertainty from sparse-observation effects, training uses fully observed spatiotemporal fields (no sensor subsampling).
For evaluation, we visualize how a massless object would move in each flow field. To track its trajectory, we recover velocities from vorticity using the classical streamfunction–vorticity formulation for 2D incompressible flow on a periodic domain, implemented via FFT-based Poisson inversion; see \citet{hussaini1986spectral,boyd2001chebyshev,kundu2024fluid} for details.

\begin{figure}[t!]
  \centering
  \begin{minipage}[t]{0.48\linewidth}
    \centering
    \includegraphics[width=\linewidth]{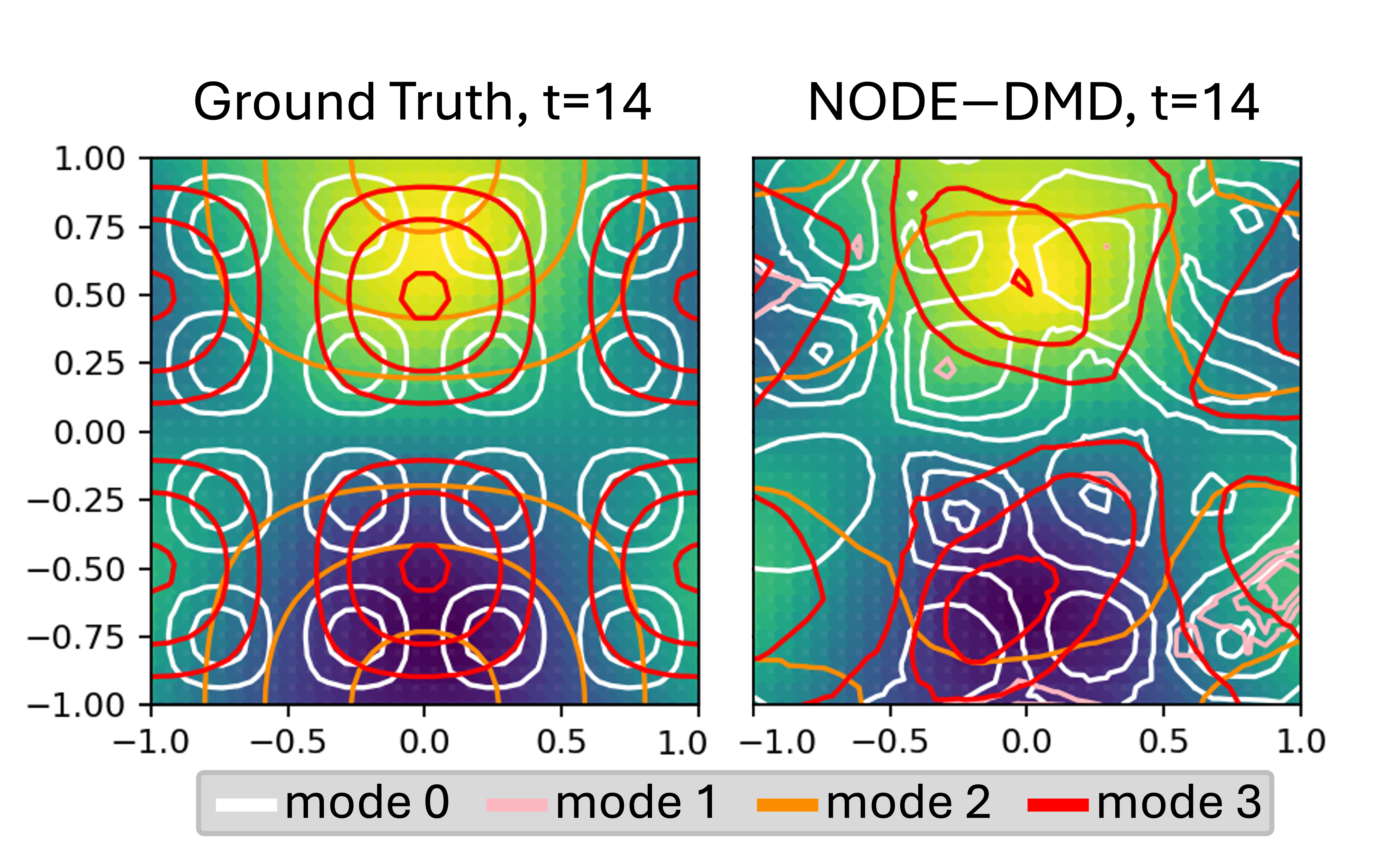}
    \captionof{figure}{\textbf{Mode-portrait overlays at absolute levels.}
      Colored contours show per-mode isocontours of $\lvert W_k(x)\,\phi_k(t)\rvert$
      at thresholds fixed once per mode from the GT sequence (30/60/90-th percentiles).
      Backgrounds show the corresponding field (Left: ground-truth mode portrait; Right: model-predicted mode portrait).}
    \label{fig:mode_contour}
  \end{minipage}\hfill
  \begin{minipage}[t]{0.48\linewidth}
    \centering
    \includegraphics[width=\linewidth]{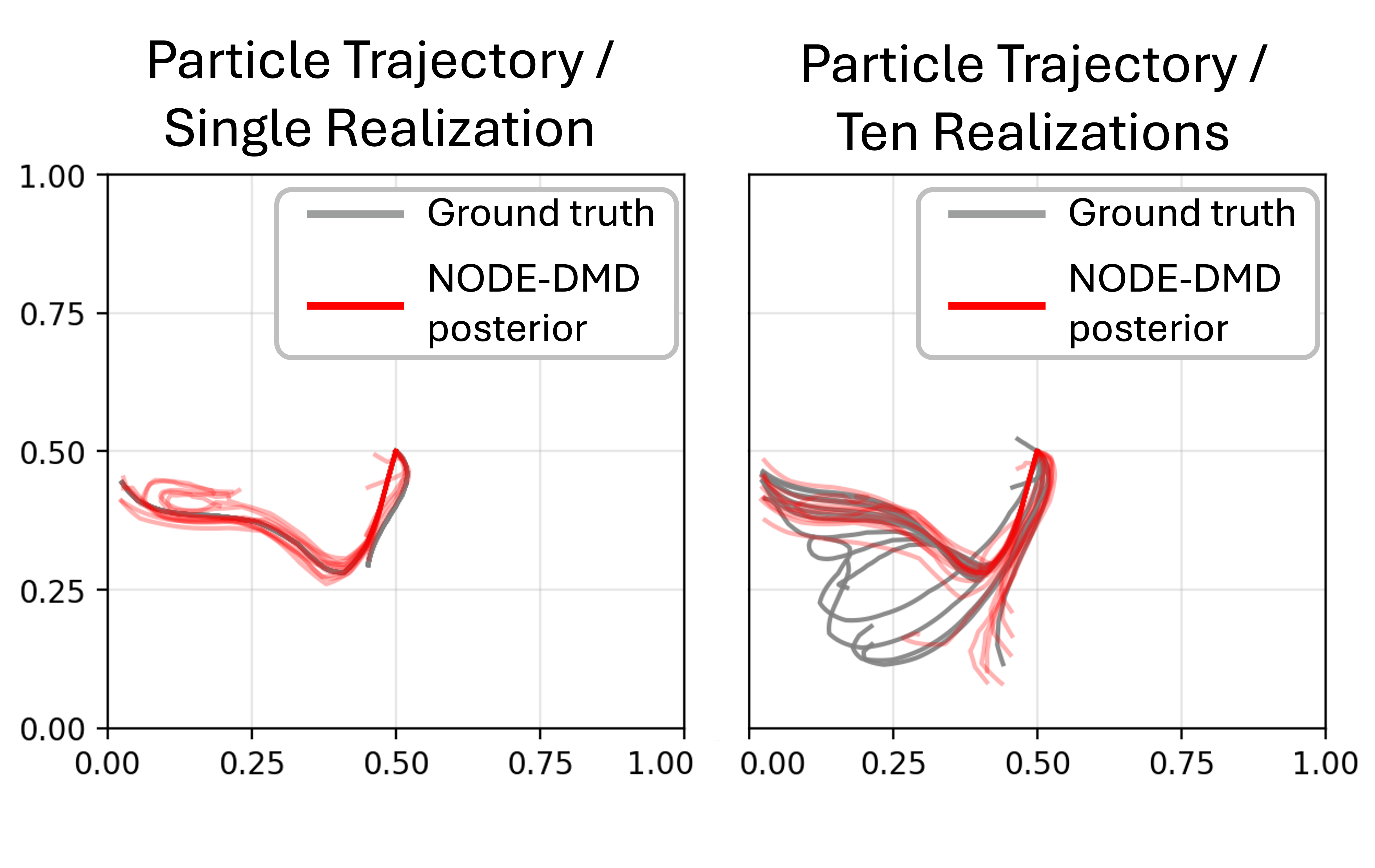}
    \captionof{figure}{\textbf{Particle trajectories from velocity field}. From posterior samples (Red) vs.\ ground truth (Gray).
      Left: Model trained on a single vorticity-field realization.
      Right: Model trained on $10$ realizations.}
    \label{fig:traj}
  \end{minipage}
\end{figure}

Figure~\ref{fig:traj} compares two training regimes. On the left, from a single realization, the posterior concentrates and the $N{=}10$ sampled trajectories cluster tightly around the mean drift, indicating limited epistemic variability when trained on one realization only. On the right, training on a dataset of $10$ realizations (same initial state but different dynamics parameters) yields a posterior that captures inter-realization variability: the ensemble exhibits a wider spread while preserving the mean drift. In other words, NODE--DMD does not collapse to an averaged, time-invariant pattern but learns a distribution that covers the family of dynamics induced by the training realizations.


\section{Conclusion}
We introduced Stochastic NODE--DMD, a probabilistic and interpretable framework for system identification that (i) reformulates DMD as a generative model, (ii) employs a neural implicit representation to enable grid-free, continuous spatial reconstruction from sparse observations, and (iii) augments the linear DMD drift with a stochastic Neural ODE to capture nonlinear residuals while propagating uncertainty. In doing so, the method preserves the spectral interpretability of DMD yet extends its applicability to sparse, noisy, and strongly nonlinear settings.

Across four benchmarks (one synthetic; reaction–diffusion, vorticity flow, cylinder flow), the method reconstructs from 10\% fixed sensors, maintains coherent geometry and phase in long-horizon rollouts, and recovers principal dynamical factors (modes and eigenvalues). Training over multiple realizations learns a calibrated distribution over dynamics rather than collapsing to time-averaged patterns. 
Our proposed method is competitive on near-linear cases and stronger on more nonlinear systems, indicating that combining linear spectral structure with learned nonlinear corrections improves robustness under sparsity and noise while retaining interpretability.

\paragraph{Limitations and future work.}
Scaling to large, real-world settings will benefit from faster latent-SDE solvers and memory-/parallel-aware training, as well as temporal regularization and physics-informed priors to further improve long-horizon stability. 
An important potential application of this framework is modeling dynamic open environment for planning and control, such as balloons and UAVs operating in wind-fields. 
Such settings often include irregular, sparse sensor networks which may be augmented through online data assimilation or active sensing.
We hope that our method for tractable, uncertainty-aware reconstruction and forecasting can interface directly with downstream planning and control in autonomous systems.


\bibliography{reference}
\newpage
\appendix
\section{Stochastic Neural ODE procedure}
\label{app:snode_details}
We evolve the latent state by the SDE in Eq.~\eqref{eq:sde}.
For implementation we work in $\mathbb{R}^{2r}$ via real–imag stacking and use the real lift $\Lambda_{\mathbb{R}}$ of $\Lambda$.
Over $\Delta t$, we apply an uncertainty-aware Euler--Maruyama with $p$ substeps $\delta t$.
See Algorithm~\ref{alg:code_snode} for the full procedure.
\begin{algorithm}[h]
\caption{Stochastic Neural ODE Integration with Uncertainty Propagation}
\label{alg:code_snode}
\KwIn{Initial latent mean \(\boldsymbol{\mu}_{\phi_t} \in \mathbb{C}^r\), covariance \(\Sigma_{\phi_t} \in \mathbb{C}^{r \times r}\), eigenvalues \(\Lambda\), neural drift \(f_\theta\), diffusion \(\tau\), time step \(\delta t\), horizon \(\Delta t\)}
\KwOut{Final latent distribution \(\mathcal{CN}(\boldsymbol{\mu}_{\phi_{t+\Delta t}}, \Sigma_{\phi_{t+\Delta t}})\)}
\For{$p = 0, 1, \dots, P-1$ \tcp*{where \(t_p = t+p \delta t\), \(\Delta t = P \delta t\)}}{
    $\boldsymbol{\delta} \gets \Lambda \boldsymbol{\mu}_{\phi_{t_p}} + f_\theta(\boldsymbol{\mu}_{\phi_{t_p}}, t_p)$\; 
    \tcp{Total drift}
    
    $\boldsymbol{\mu}_{\phi_{t_{p+1}}} \gets \boldsymbol{\mu}_{\phi_{t_p}} + \Delta t \cdot \boldsymbol{\delta}$\; 
    \tcp{Mean update}
    
    $J \gets \nabla_{\boldsymbol{\phi}} (\Lambda \boldsymbol{\phi} + f_\theta(\boldsymbol{\phi}, t_p)) \big|_{\boldsymbol{\phi} = \boldsymbol{\mu}_{\phi_{t_p}}})$; 
    \tcp{Jacobian of drift}
    
    \(A \gets I + \Delta t \cdot J\)\;
    
    \(\Sigma_{\phi_{t_{p+1}}} \gets A \Sigma_{\phi_{t_p}} A^H + \Delta t \cdot \tau^2 I\)\; 
    \tcp{Covariance propagation}
}
\Return{\(\mathcal{CN}(\boldsymbol{\mu}_{\phi_{t+\Delta t}}, \Sigma_{\phi_{t+\Delta t}})\)}
\end{algorithm}

\section{Training Loss Function}
\label{app:loss}
For a transition from time \(t_k\) to \(t_{k+1}\), let \(\mathbf{y}_{k+1} \in \mathbb{C}^m\) be the target observation, \((\boldsymbol{\hat\mu}_y, \boldsymbol{\hat\sigma}_y^2)\) the predicted mean and variance of the reconstructed field, and \((\boldsymbol{\mu}_{\phi}, \boldsymbol{\sigma}_{\phi}^2)\), \((\boldsymbol{\hat\mu}_{{\phi}}, \boldsymbol{\hat\sigma}_{{\phi}}^2)\) the latent distributions from the \texttt{Latent Encoder} and Stochastic \texttt{Neural ODE}, respectively.

The total loss is
\begin{equation}
\label{eq:loss_app}
\mathcal{L} = w_{\text{recon}} \cdot \mathcal{L}_{\text{recon}} + w_{\text{kl}} \cdot \mathcal{L}_{\text{kl}} + w_{\text{cons}} \cdot \mathcal{L}_{\text{cons}},
\end{equation}
with weights \(w_{\text{recon}} = 3\), \(w_{\text{kl}} = 10^{-3}\), \(w_{\text{cons}} = 0.15\).

\begin{itemize}
    \item \textbf{Reconstruction Loss}: A Gaussian negative log-likelihood encourages accurate field prediction while adapting to data uncertainty:
      \[
      \mathcal{L}_{\text{recon}} = \frac{1}{2} \left[ \log \boldsymbol{\hat\sigma}_y^2 + \frac{(\mathbf{y}_{k+1} - \boldsymbol{\hat\mu}_y)^2}{\exp(\log \boldsymbol{\hat\sigma}_y^2)} + \log(2\pi) \right].
      \]
    \item \textbf{Latent KL Regularization}: A KL divergence term regularizes the latent mode coefficients \(\boldsymbol{\hat\phi}_i\) toward a standard complex Gaussian prior, preventing overfitting and promoting structured representations:
      \[
      \mathcal{L}_{\text{kl}} = -\frac{1}{2} \sum_i \left( 1 + \log \boldsymbol{\hat\sigma}_{\phi}^2 - \boldsymbol{\hat\mu}_{\phi}^2 - \exp(\log \boldsymbol{\hat\sigma}_{\phi}^2) \right).
      \]
    \item \textbf{Consistency Loss}: Latent distribution matching loss aligns the encoder-estimated distribution \(p(\boldsymbol{\phi}_{k+1} \mid \mathbf{y}_{k+1}, \gamma(S))\) with the SDE-propagated distribution \(p(\boldsymbol{\hat{\phi}}_{k+1} \mid \boldsymbol{\phi}_{k})\) to enforce temporal consistency:
      \[
      \mathcal{L}_{\text{cons}} = \text{MSE}(\boldsymbol{\mu}_{\phi}, \boldsymbol{\hat\mu}_{{\phi}}) + \kappa \cdot \text{KL}\bigl( \mathcal{CN}(\boldsymbol{\mu}_{\phi}, \sigma^2_{\phi}) \,\|\, \mathcal{CN}(\boldsymbol{\hat\mu}_{{\phi}}, \hat\sigma^2_{{\phi}}) \bigr),
      \]
      where \(\kappa = 0.001\) scales the variance term.
\end{itemize}
\section{Training Detail}
\label{app:train_detail}
The model is trained end-to-end using gradient descent on the training dataset. Due to its recursive structure, we can select either ground-truth observations $\mathbf{y}_k$ or model predictions $\mathbf{\hat{y}}_k$ as inputs during training. To stabilize the process and enable robust multi-step forecasting, we apply a curriculum learning schedule that transitions from teacher forcing (using $\mathbf{y}_k$) to autoregressive mode (using $\mathbf{\hat{y}}_k$). For each training batch, we uniformly select teacher forcing mode with probability $\epsilon$ {applying it to all sequences and timesteps in the batch) or autoregressive mode with probability $1 - \epsilon$, where $\epsilon$ decays linearly from 1 to 0 over the course of training. In fully autoregressive mode, the model reconstructs the entire sequence of duration $T$ given only the initial observation.
\section{Simulation Dataset Details}
\label{app:simulations}

\subsection{Synthetic DMD Sequence}
We construct a synthetic dataset based on the principles of DMD, where the underlying dynamics are generated from a known set of modes, eigenvalues, and initial coefficients. This allows direct comparison between learned Koopman modes and ground-truth DMD components. The state at each time step is defined as a linear combination of spatial modes $W(x)$ modulated by complex exponentials. Let
\[
\begin{aligned}
m_0(x,y) &= \sin\!\Big(\tfrac{\pi}{2}(x{+}1)\Big)\,\cos\!\Big(\tfrac{\pi}{2}(y{+}1)\Big),\\
m_1(x,y) &= \cos\!\big(\pi(x{+}1)\big)\,\sin\!\big(\pi(y{+}1)\big),\\
m_2(x,y) &= \sin(2\pi x)\,\sin(2\pi y),\\
m_3(x,y) &= 0.5,
\end{aligned}
\]
and define \(W(x)=\big[m_0(x),m_1(x),m_2(x),m_3(x)\big]^\top\).
The field is a linear combination
\[
I(x,t)\;=\;W(x)^\top \,\phi(t),\qquad 
\phi_k(t)\;=\;b_k \exp\!\big((\alpha_k + j\omega_k)\, t\,\Delta t_{\mathrm{eff}}\big),
\]
with \(\Delta t_{\mathrm{eff}}=0.1\) (implemented via a fixed time-scaling so that sequences share identical temporal factors across experiments).

We use
\[
\alpha=[-0.01,\,-0.05,\,-0.20,\,-0.01],\quad
\omega=[2.00,\,4.00,\,1.00,\,0.30],\quad 
\]
\[
b=[1.0{+}0.5j,\ 0.8{-}0.3j,\ 0.7{+}0.2j,\ 0.2{+}0.0j].
\]

The domain is a uniform \(32{\times}32\) grid over \([{-}1,1]\times[{-}1,1]\) (\(n{=}1024\)). At each time \(t\), only \(10\%\) of grid points (a single, time-invariant random index set) are observed to emulate sparse sensing. Observations are corrupted by i.i.d. complex Gaussian noise \( \eta\sim\mathcal{N}_\mathbb{C}(0,\sigma^2) \) with \(\sigma=0.1\):
\[
y_t(x_i)\;=\;I(x_i,t)+\eta_{i,t}.
\]
We roll out \(T{=}100\) steps with timestamps \(t=0,\dots,T{-}1\). For evaluation, we also retain the full noiseless fields \(I(x,t)\). The ground-truth discrete-time eigenvalues used for comparison are
\[
\mu_k=\exp\!\big((\alpha_k + j\omega_k)\,\Delta t_{\mathrm{eff}}\big).
\]
\subsection{Gray-Scott Reaction-Diffusion Model}

The {Gray-Scott model} simulates the interaction of two chemical species $u$ and $v$ through reaction, diffusion, and feed/kill processes, exhibiting rich pattern formation such as spots, waves, and mazes. The evolution is governed by the coupled PDEs:
\begin{align}
    \frac{\partial u}{\partial t} &= D_u \nabla^2 u - u v^2 + F(1 - u), \\
    \frac{\partial v}{\partial t} &= D_v \nabla^2 v + u v^2 - (F + k) v,
\end{align}
with periodic boundary conditions. The Laplacian is approximated using finite differences with nearest-neighbor rolls. We use parameters $D_u = 2 \times 10^{-4}$, $D_v = 1 \times 10^{-5}$, $F = 0.035$, $k = 0.065$, known to produce \textbf{wave-like and spotted patterns}~\cite{pearson1993complex}. A small perturbation is added to $F$ at initialization ($\sigma = 10^{-3}$) to break symmetry.

The system is simulated on a {$100 \times 100$ grid} over $[-1, 1] \times [-1, 1]$, with spatial step $\Delta x = 0.01$. Initial conditions are wave-modulated fields:
\begin{align}
    u(x,y,0) &= 0.9 + 0.1 \sin(4\pi x) \cos(2\pi y), &
    v(x,y,0) &= 0.1 + 0.05 \sin(\pi x).
\end{align}
Only {10\% of spatial points} are observed. The concentration of species $v$ is used as the observable.

\subsubsection{2D Navier-Stokes Vorticity Flow (Spectral Method)}

We simulate incompressible 2D fluid flow using the {Navier-Stokes equations} in vorticity-stream function form, using~\cite{cao2024spectral}. The vorticity $\omega = \nabla \times \mathbf{u}$ evolves as:
\begin{equation}
    \frac{\partial \omega}{\partial t} + (\mathbf{u} \cdot \nabla)\omega = \nu \nabla^2 \omega,
\end{equation}
with $\mathbf{u} = \nabla^\perp \psi$ and $\nabla^2 \psi = -\omega$. The initial vorticity field is a filtered random field with peak wavenumber $k = 2$, ensuring smooth, coherent structures. The simulation uses the \texttt{torch-cfd} library with {Crank-Nicolson + RK4 time stepping}, viscosity $\nu = 10^{-3}$, and domain $[0, 2\pi] \times [0, 2\pi]$.

The grid resolution is \textbf{$128 \times 128$} (coordinates normalized to $[-1, 1] \times [-1, 1]$). Time step is $\Delta t = 10^{-3}$, with snapshots saved every 100 steps ($\sim \Delta t_{\text{save}} = 0.1$), yielding 100 snapshots over $T = 10.0$. The flow develops {vortex streets and turbulent-like structures}.

\subsubsection{2D Navier-Stokes Flow Past a Cylinder}

We simulate {viscous incompressible flow around a circular cylinder} using~\cite{cao2024spectral}, modeling the {von K\'arm\'an vortex street}. The Navier-Stokes equations are:
\begin{align}
    \frac{\partial \mathbf{u}}{\partial t} + (\mathbf{u} \cdot \nabla)\mathbf{u} &= -\nabla p + \nu \nabla^2 \mathbf{u}, \\
    \nabla \cdot \mathbf{u} &= 0,
\end{align}

The grid is \textbf{$100 \times 100$} in $[-1, 1] \times [-1, 1]$. Viscosity is $\nu = 1/500$, time step $\Delta t = 10^{-3}$, and snapshots are saved every 150 steps over $T = 2.0$. The vorticity field exhibits {periodic shedding} behind the cylinder. Spatial observations are densely available (full cropped grid), but only {vorticity} is used as the observable. 

\section{Recovering Ground-Truth Dynamics from Noisy, Sparse Observations}
\label{app:dyanmics_retriever}
We evaluate the model trained in Sec.~\ref{sec:gt-dynamics} and report (i) spatial mode similarity and (ii) continuous-time eigenvalue accuracy.  
\subsection{Metric details}
\paragraph{Mode similarity} ($\mathrm{sim}_{\cos}(\widehat W, W_{\text{GT}})$).
Learned modes $\widehat W\in\mathbb{C}^{N\times r}$ are matched to ground-truth modes $W\in\mathbb{C}^{N\times r}$ by solving a linear assignment (Hungarian) problem over permutations $\pi$ that maximizes per-mode cosine similarity after a phase correction $e^{i\theta_k}$:
\[
\mathrm{cos}(k)\;=\;\frac{\langle \widehat W_{:,k},\,e^{i\theta_k}W_{:,\pi(k)}\rangle}
{\|\widehat W_{:,k}\|\,\|W_{:,\pi(k)}\|}\in[0,1].
\]
We report the mean per-pair cosine similarity.

\paragraph{Eigenvalue accuracy} ($|\widehat\lambda-\lambda_{\text{GT}}|$).
Pairs $(\widehat\lambda_i,\lambda_j)$ follow the mode assignment.
Because $\phi_k(t)$ evolves under a Neural ODE, we estimate continuous-time eigenvalues using a log-ratio estimator~\cite{tu2014exact,schmid2010dmd,brunton2016book}:
\[
\widehat\lambda_k(t)\;=\;\frac{1}{\Delta t}\,\operatorname{Log}\!\Big(\frac{\phi_k(t+\Delta t)}{\phi_k(t)}\Big),
\quad
\widehat\lambda_k\;=\;\mathrm{median}_{t}\,\widehat\lambda_k(t),
\]
with a time-consistent branch choice. We then report the mean absolute complex difference.

 \subsection{Quantitative results.}
Table~\ref{tab:gt-dynamics} compares the proposed NODE--DMD and NDMD on mode agreement and spectral accuracy.

\textbf{Mode similarity.} Our average cosine is $0.708$ (per-mode $0.589,\,0.811,\,0.628,\,0.807$), while NDMD records $0.767$ on average, reflecting very high alignment on three modes ($0.999,\,0.999,\,0.996$) and a lower value on one mode ($0.074$).
Overall, NDMD attains a slightly higher per-mode cosine, whereas our method exhibits a more even alignment across modes.
Since the cosine is computed per mode and is sensitive to basis rotations within a shared subspace, these differences do not necessarily imply a gap in reconstruction quality.

\textbf{Eigenvalue accuracy.} Our average absolute complex error is \textbf{0.607}, compared to \textbf{1.775} for NDMD.
NDMD’s estimates are closer to low/near-zero frequencies, yielding larger differences from the ground-truth spectrum. 
By contrast, our estimates preserve the overall ordering and scales of damping and frequency.
Given that reconstructions follow $\mathbf{y}(t)=W\,\phi(t)$ and long-horizon behavior is largely governed by spectral factors in $\phi(t)$ (phase and amplitude growth), these results suggest that, despite a slightly lower per-mode cosine, the improved spectral estimates of our method are particularly beneficial for stable, long-horizon evolution.

\textbf{Side note (fair comparison with NDMD).}
Neural DMD (NDMD) parameterizes one DC mode with eigenvalue $0{+}0j$ and $r/2$ complex-conjugate pairs. 
For real-field reconstruction it uses only the positive member with a factor $2\,\mathrm{Re}(\cdot)$ (since $z+z^*=2\,\mathrm{Re}\,z$).
For a like-for-like comparison, we evaluate NDMD using the single DC mode and the three \emph{positive} members of the conjugate pairs (the $2\,\mathrm{Re}(\cdot)$ form is equivalent to summing both members).
Eigenvalue metrics are computed on the positive-branch eigenvalues accordingly.
We also note that the ground-truth decomposition includes a spatial DC component, which aligns well with NDMD’s dedicated DC mode; thus this setting is, if anything, slightly favorable to NDMD.
Despite these architectural differences, this protocol is intended to keep the comparison as fair as possible.
\begin{table}[t]
\centering
\small
\begin{tabular}{lccccc}
\hline
Metric & mode 0 & mode 1 & mode 2 & mode 3 & Avg. \\
\hline
\hline
$\mathrm{sim}_{\cos}(\hat{W}, W_{\text{GT}})_\text{NODE-DMD}$ & 0.589 & 0.811 & 0.628 & 0.807 & \textbf{0.708} \\
$\mathrm{sim}_{\cos}(\hat{W}, W_{\text{GT}})_\text{NDMD}$ & 0.0741 & 0.999 & 0.999 & 0.996 & \textbf{0.7674} \\
$|\widehat{\lambda}-\lambda_{\text{GT}}|_\text{NODE-DMD}$ & 0.060 & 0.985 & 1.248 & 0.135 & \textbf{0.607} \\
$|\widehat{\lambda}-\lambda_{\text{GT}}|_\text{NDMD}$ & 0.266 & 1.0198 & 1.937 & 3.8754 & \textbf{1.7748} \\
\hline
$\lambda_{\text{GT}}$ & $-0.01{+}2.0j$ & $-0.05{+}4.0j$ & $-0.20{+}1.0j$ & $-0.01{+}0.3j$ &  \\
$\widehat{\lambda}_\text{NODE-DMD}$ & $-0.22{+}1.03j$ & $-0.03{+}2.75j$ & $-0.25{+}0.88j$ & $0.05{+}0.32j$ &  \\
$\widehat{\lambda}_\text{NDMD}$ & $-0.002{+}0.033j$ & $0{+}0j$ & $-0{+}0.06j$ & $0.002{+}0.125j$ &  \\
\hline
\end{tabular}
\caption{\textbf{Mode agreement and continuous-time eigenvalue errors.} 
Cosine similarity is computed after permutation and per-mode phase alignment.
Eigenvalue errors are absolute complex differences after optimal assignment.}
\label{tab:gt-dynamics}
\end{table}

\section{Resolution flexibility Experiment}
A key advantage of NODE–DMD is its grid-free formulation, which allows the model to reconstruct the full spatial field at arbitrary position without requiring retraining. To evaluate this, we compare NODE–DMD with NDMD~\cite{saraertoosi2025neural} under low- and high-resolution reconstruction settings. For each task, we downsample the field to a $(50 \times 50)$ grid for the low-resolution experiment and upsample it to a $(200 \times 200)$ grid for the high-resolution experiment. All models are trained only on the original resolution data with 10\% spatial sampling.
\begin{figure}
    \centering
    \includegraphics[width=1\linewidth]{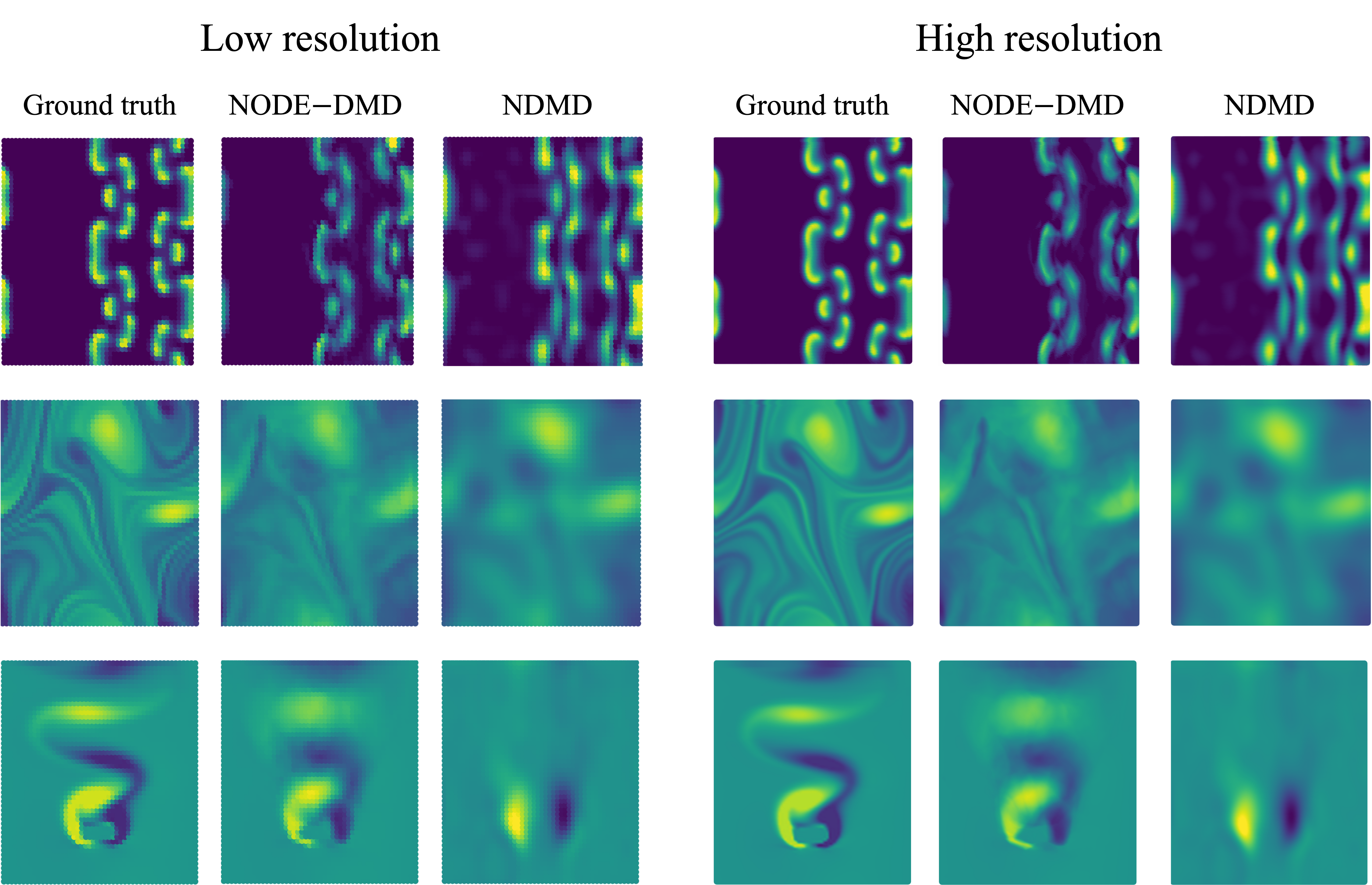}
    \caption{Reconstruction results at the final time step for low-resolution (50$\times$50) and high-resolution (200$\times$200) grids for the Gray--Scott system (first row), Vorticity flow (second row), and Cylinder flow (third row). Visualizations are generated using NDMD and the 1-step NODE--DMD model.}
    \label{fig:resolution}
\end{figure}

\begin{table}
\footnotesize
\centering
\begin{tabular}{c|ccc|ccc}
\multirow{1}{*}{} &
\multicolumn{3}{c|}{\textsc{Low resolution}} &
\multicolumn{3}{c}{\textsc{High resolution}} \\

\hline
{Method} & NDMD & \makecell{NODE\\DMD(1)} & \makecell{NODE\\DMD(m)}
         & NDMD & \makecell{NODE\\DMD(1)} & \makecell{NODE\\DMD(m)} \\ 
\hline

{\textsc{Gray--Scott}}
& 0.0048{\tiny ($-$2.0\%)} 
& 0.0047{\tiny (+2.2\%)} 
& 0.0078{\tiny (+2.6\%)} 
& 0.0047{\tiny ($-$4.1\%)} 
& 0.0045{\tiny ($-$2.2\%)} 
& 0.0075{\tiny ($-$1.3\%)} \\

{\textsc{Vorticity Flow}}
& 0.0047{\tiny (0\%)} 
& 0.0024{\tiny (0\%)} 
& 0.0024{\tiny (0\%)} 
& 0.0047{\tiny (0\%)} 
& 0.0024{\tiny (0\%)} 
& 0.0024{\tiny (0\%)} \\

{\textsc{Cylinder Flow}}
& 0.0507{\tiny (+8.6\%)} 
& 0.0064{\tiny ($-$4.5\%)} 
& 0.0462{\tiny ($-$2.3\%)} 
& 0.0530{\tiny (+13.5\%)} 
& 0.0067{\tiny (0\%)} 
& 0.0477{\tiny (+0.8\%)} \\
\hline

\end{tabular}
\caption{$L_1$ reconstruction error and its relative change under different spatial resolutions. NODE--DMD(1) and NODE--DMD(m) denote 1-step and $m$-step prediction variants, respectively. Percentage change is computed with respect to the original-resolution full-field reconstruction error reported in Table~\ref{tab:reconstructions}, where negative values indicate error reduction and positive values indicate error increase.}\label{tab:resolution}
\end{table}
Figure~\ref{fig:resolution} presents qualitative reconstruction results at the final time step for the three benchmark systems. Table~\ref{tab:resolution} reports the average per-pixel $L_1 $error over the full temporal rollout for low and high resolutions, along with the percentage change relative to the original-resolution reconstruction error reported in Table~\ref{tab:reconstructions}. Negative percentages indicate error reduction, and positive values indicate error increase.

Across all tasks, NODE–DMD exhibits consistently small positive changes in reconstruction error when the target resolution differs from the training resolution compared to NDMD. These results highlight the robustness of NODE–DMD to changes in grid density and demonstrate that its continuous implicit representation enables stable full-field reconstruction, independent of the resolution at which the field is queried.


\end{document}